\documentclass[letterpaper, 10 pt, onecolumn]{ieeeconf}  

\IEEEoverridecommandlockouts                              

\usepackage{titlesec}
\usepackage{lipsum}
\usepackage{cite}
\usepackage[textwidth=1.4cm,textsize=scriptsize]{todonotes}
\usepackage[noend]{algpseudocode}
\usepackage{subfig}
\usepackage{mathtools}
\usepackage{bm}
\usepackage{algorithmicx}
\usepackage{algorithm}
\usepackage{algpseudocode}
\usepackage{graphicx}
\usepackage{epstopdf}
\usepackage{amsfonts}
\usepackage{amsmath}
\usepackage{dsfont}

\usepackage{epsfig} 
\usepackage{xcolor}
\usepackage{amssymb}

\usepackage{cuted}
\let\labelindent\relax
\usepackage{enumitem}
\usepackage{tikz}
\usepackage{balance}
\usepackage{soul}
\usepackage[normalem]{ulem}

\usetikzlibrary{shapes}
\usetikzlibrary{positioning}

\usepackage[hidelinks]{hyperref}
\def \strategyname/{\textsc{MRoPE}}
\title{\strategyname/: A \underline{M}ulti-\underline{Ro}bot Safe Cooperative Strategy via combined\\ \underline{P}redictive Safety Filters and \underline{E}llipse-based Constraint Compression

\thanks{Work partially funded by the European Union - Next Generation EU - under the National Recovery and Resilience Plan (NRRP), Mission 4, Component
2, Investment 3.3; CUP J33C23001440009 and by Leonardo Innovation Labs.}

\thanks{$^1$ are with the Department of Electrical, Electronic and Information Engineering, Alma Mater Studiorum - Universit\`a di Bologna, Bologna, Italy. 
        {\tt\small \{a.rosetti, lorenzo.pichierri, giuseppe.notarstefano\}@unibo.it}}

\thanks{$^2$ are with Leonardo S.p.a., Leonardo Innovation Labs, Rome, Italy. 
        {\tt\small \{domenico.cappello, fabrizio.schiano\}@leonardo.com}}
}
\author{A.~Rosetti$^1$, L.~Pichierri$^1$, D.~Cappello$^2$, F.~Schiano$^2$, and G. Notarstefano$^1$}%
\date{\today}

\def\er/{Erd\H{o}s-R\'enyi}
\DeclareMathOperator*{\argmax}{arg\,max}
\newcommand{\ti}{t}
\newcommand{\dt}{\Delta\ti}
\newcommand{\iter}{\tau}
\newcommand{\Hor}{T}
\newcommand{\N}{N}
\newcommand{\Ag}{\mathcal{V}}
\newcommand{\Nobs}{N_{O}}
\newcommand{\real}{\mathbb{R}}
\newcommand{\rag}{r_{\text{dr}}}
\newcommand{\robs}{r_{j,\text{obs}}}
\newcommand{\sdist}{r_{\text{sens}}}

\newcommand{\sdim}{n}
\newcommand{\udim}{m}
\newcommand{\pdim}{d}

\newcommand{\xa}{z}
\newcommand{\xai}{\xa_{i}}
\newcommand{\xat}{\xa^{\ti}}
\newcommand{\xatp}{\xa^{\ti+1}}
\newcommand{\xati}{\xat_{i}}
\newcommand{\xatpi}{\xatp_{i}}
\newcommand{\ptides}{\xatpi}
\newcommand{\proj}{\mathrm{P}}
\newcommand{\projell}{\proj_{\ellipsesmall}}
\newcommand{\xproj}{\projell(\ptides)}

\newcommand{\da}{d_{b}}

\newcommand{\state}{x}
\newcommand{\statet}{\state^{\ti}}
\newcommand{\stateti}{\statet_{i}}
\newcommand{\statetpi}{\state^{\ti+1}_{i}}
\newcommand{\statebfi}{{\state_{i}}}
\newcommand{\stateki}{\state^{\iter}_{i}}
\newcommand{\statekpi}{\state^{\iter+1}_{i}}

\newcommand{\p}{p}
\newcommand{\pii}{\p_{i}}
\newcommand{\pt}{\p^{\ti}}
\newcommand{\pti}{\pt_{i}}
\newcommand{\ptihat}{\hat{\p}^{\ti}_{i}}
\newcommand{\perr}{\tilde{\p}^{\ti}_{i}}
\newcommand{\ptj}{\pt_{j}}
\newcommand{\pki}{p^{\iter}_{i}}

\newcommand{\slack}{\varepsilon}
\newcommand{\slacki}{\slack_{i}}

\newcommand{\slackj}{\slack_{ij}}

\newcommand{\slackkj}{\slack^{\iter}_{ij}}

\newcommand{\uu}{u}
\newcommand{\inputti}{\uu^{\ti}_{i}}
\newcommand{\inputtides}{\uu^{\ti}_{i,\text{des}}}
\newcommand{\inputki}{\uu^{\iter}_{i}}
\newcommand{\inputbfi}{{\uu_{i}}}
\newcommand{\inputmin}{\uu_{\text{min}}}
\newcommand{\inputmax}{\uu_{\text{max}}}

\newcommand{\vel}{v}

\newcommand{\velki}{\vel^{\iter}_{i}}
\newcommand{\veltihat}{\hat{\vel}^{\ti}_{i}}
\newcommand{\velerr}{\tilde{\vel}^{\ti}_{i}}
\newcommand{\velmin}{\vel_{\text{min}}}
\newcommand{\velmax}{\vel_{\text{max}}}
\newcommand{\veldes}{\vel_i^{\star,\ti}}

\newcommand{\obs}{o}
\newcommand{\obsj}{\obs_{j}}

\newcommand{\targett}{\target^{\ti}}
\newcommand{\target}{c}
\newcommand{\targetvel}{v_\target}
\newcommand{\targetk}{\target^{\iter}}
\newcommand{\targetkp}{\target^{\iter+1}}
\newcommand{\ellipse}{\mathcal{E}}
\newcommand{\ellipset}{\bar{\ellipse}^{\ti}_i}
\newcommand{\ellipsetb}{{\ellipse}^{\ti}_i}
\newcommand{\ellipsesmall}{\bar{\ellipse}^{\ti}_i}
\newcommand{\Mmat}{M}
\newcommand{\Mmatt}{\Mmat^{\ti}_{\text{nom},i}}
\newcommand{\Mmattb}{{\Mmat}^{\ti}_i}
\newcommand{\Mmatkb}{{\Mmat}^{\iter}_i}

\newcommand{\cell}{c_{\text{ell}}}

\newcommand{\cellk}{\cell^{\iter}}

\newcommand{\zh}{w}
\newcommand{\ztj}{\zh^{\ti}_{ij}}
\newcommand{\zkj}{\zh^{\iter}_{ij}}
\newcommand{\zkpj}{\zh^{\iter+1}_{ij}}
\newcommand{\mtj}{m^{\ti}_{ij}}
\newcommand{\mkj}{m^{\iter}_{ij}}
\newcommand{\mkpj}{m^{\iter+1}_{ij}}
\newcommand{\Tk}{T^{\iter+1}_{\iter}}

\newcommand{\costati}{\costat_{i}}
\newcommand{\costat}{\costa^{\ti}}
\newcommand{\costa}{\ell}
\newcommand{\costatpi}{\costa^{\ti+1}_{i}}

\newcommand{\costsuti}{J^{\ti}_{u,i}}

\DeclareMathOperator{\col}{\text{col}}
\newcommand{\until}[1]{\{1,\ldots,#1\}}
\newcommand{\tHor}{\{\ti,\ldots,\ti + \Hor\}}

\newcommand{\Neighs}{\mathcal{N}^{\text{s},\ti}_{i}}
\newcommand{\Neigha}{\mathcal{N}^{\text{a}}_{i}}

\newcommand{\adj}{\mathcal{A}}

\newcommand{\activeset}{\mathcal{S}}
\newcommand{\activeseti}{\activeset_{i}}
\newcommand{\constrexc}{\mathcal{C}}

\newcommand{\tBki}{\tilde{B}_{i}^{k}}
\newcommand{\tcellki}{\tilde{c}_{\text{ell}, i}^{k}}
\newcommand{\tactseti}{\tilde{\activeset}_{i}}
\newcommand{\tactsetki}{\tactseti^{k}}
\newcommand{\tconstrexcki}{\tilde{\constrexc}_{i}^{k}}
\newcommand{\tconstrexckj}{\tilde{\constrexc}_{j}^{k}}
\newcommand{\celli}{c_{\text{ell}, i}}

\newcommand{\cellti}{\celli^{\ti}}

\newcommand{\Bti}{B_{i}^{\ti}}

\newtheorem{theorem}{Theorem}[section]

\newtheorem{definition}[theorem]{Definition}

\newtheorem{remark}[theorem]{Remark}

\newtheorem{problem}[theorem]{Problem}

\newcommand\oprocendsymbol{\hbox{$\square$}}
\newcommand\oprocend{\relax\ifmmode\else\unskip\hfill\fi\oprocendsymbol}

\begin{document}

\maketitle
\begin{abstract}
Deploying drone swarms to track a dynamic target in cluttered environments presents severe 
computational and safety challenges. We propose \strategyname/, a hierarchical strategy that decouples the 
cooperative monitoring mission from strict local safety requirements. To overcome the computational bottlenecks 
typical of dense spaces, our approach dynamically aggregates complex obstacle geometries into a single safe bounding 
ellipse for each drone. Methodologically, this architecture is realized by combining distributed aggregative optimization 
for high-level swarm coordination, a decentralized consensus scheme for the safe area computation, and local Predictive 
Safety Filters (PSF) for real-time collision avoidance. Virtual and real-world experiments validate the framework, 
demonstrating superior real-time efficiency and scalability compared to centralized approaches. \\

Video: \url{https://youtu.be/e9WTZL23ICY}
\end{abstract}
\section{Introduction}
The use of multi-robot teams to cooperatively accomplish complex
tasks enhances efficiency and robustness across a wide range of
applications~\cite{gao22swarm}, such as
surveillance~\cite{saska2016swarm} and
mapping~\cite{wang2023distributed}. In this work, we investigate a
scenario in which a team of drones wants to safely monitor and
follow a moving target while avoiding both inter-agent collisions
and collisions with obstacles. 

A powerful tool to address
complex tasks in multi-robot applications is distributed
optimization, see,
e.g.,~\cite{shorinwa2024distributed,testa2025tutorial}.
Recently, distributed aggregative
optimization has gained
attention~\cite{li2021distributed2,carnevale2022distributed}
and has been used for modeling cooperative multi-robot
scenarios~\cite{pichierri2026multi,huang2025distributed}.
However, to inherently
ensure collision avoidance among robots or with obstacles, local
and coupling constraints must be added into the optimization
problem. This significantly increases the computation complexity,
especially in presence of a large number of obstacles.

To address safety, several approaches have been proposed.
Buffered Voronoi Cells are introduced in~\cite{zhou2017fast} to
ensure safety in a distributed fashion using only position
information. In~\cite{vinod2024decentralized}, a decentralized
Reinforcement Learning control strategy is combined with
Buffered Voronoi Cells. Integration of a distributed Model
Predictive Control (MPC) with an on-demand collision avoidance
method is proposed in~\cite{luis2020online} and it is implemented in
a parallel fashion. In~\cite{soria2021distributed} a distributed
MPC enforces also safety constraints using neighbors' planned
trajectories. Another decentralized approach is presented
in~\cite{toumieh2022decentralized} where planned optimal
trajectories of the other agents are used to create Time-Aware Safe
Corridors.
\begin{figure}[t]
	\centering
	\includegraphics[width=0.7\columnwidth]{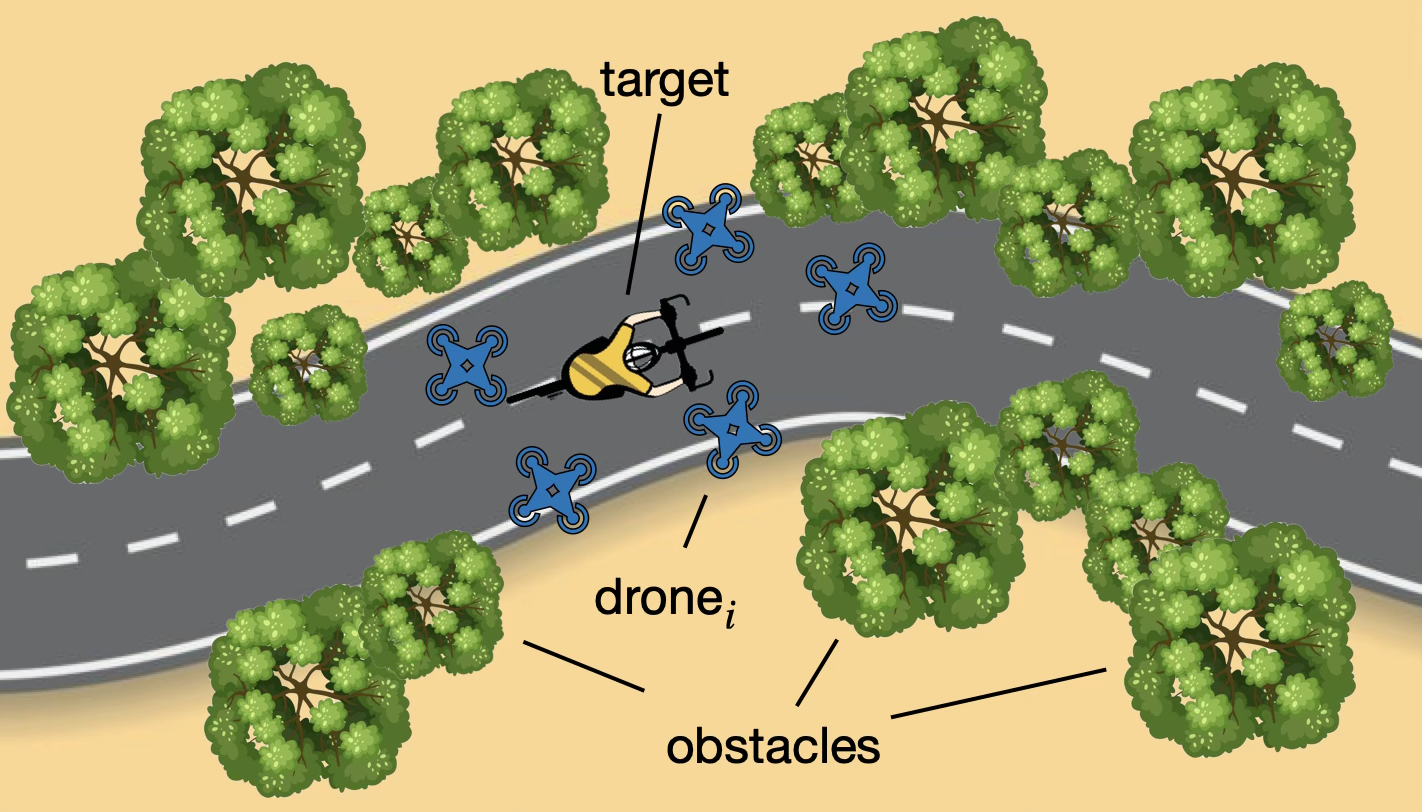}
	\caption{
	    Concept picture of the addressed scenario.
	}
	\label{fig:problem_setup}
\end{figure}

Most existing collision avoidance approaches struggle to scale with the number of agents and obstacles, 
or tightly couple safety constraints with the primary robotic task. 
To effectively decouple these objectives, Safety Filters represent a highly effective paradigm. 
In this context, Control Barrier Functions (CBFs) \cite{cavorsi2023multirobot,hu2022decentralized,cheng2020safe} 
have gained significant traction due to their minimal computational overhead in multi-robot scenarios. 
However, being purely reactive, CBF-based methods are prone to local deadlocks
in cluttered environments and often struggle to systematically handle strict actuation limits \cite{grover2020deadlock}. 
To overcome these limitations, Predictive Safety Filters (PSFs) stand out as a robust alternative.
An early work is~\cite{wabersich2018linear}, where PSF 
based on MPC techniques are used to ensure safety along a trajectory. 
However, these filters have been mainly limited to single-robot applications, 
such as racing cars~\cite{tearle2021predictive}, quadrotors~\cite{bejarano2023multi}, 
or marine navigation~\cite{vaaler2024modular}. 
While~\cite{safaoui2024safe} considers a multi-agent scenario, it relies on a 
centralized PSF. 
Multi-robot scenarios call for a proper decentralized extension of these 
filtering mechanisms, fully integrated into a high-level distributed control strategy.

The contributions of this paper are twofold. 
First, we decouple the high-level cooperative surveillance task (based on~\cite{carnevale2022distributed}) 
from the low-level safety constraints. Unlike existing works that tightly couple these aspects, 
we introduce a novel decentralized PSF relying on a low-complexity MPC.
This filter enforces both inter-agent and static obstacle collision avoidance while minimally deviating 
from the optimal reference trajectory. 
Second, to overcome computational bottlenecks in dense environments, 
we compress linear polyhedral constraints into a single ellipsoidal constraint. 
This geometric compression ensures the downstream local MPC complexity remains strictly bounded and 
independent of obstacle density. Furthermore, the ellipse adapts its eccentricity to the available free space, forcing the multi-agent team to adapt its topology and safely traverse tight spaces (e.g., in single-line formations) 
without requiring ad-hoc reconfiguration rules.

The paper unfolds
as follows. In Section~\ref{sec:problem} we introduce the
problem setup. In Section~\ref{sec:control} we detail the
hierarchically-distributed control strategy.
Section~\ref{sec:implementation} provides virtual experiments and Monte Carlo campaigns, whereas Section~\ref{sec:experiments} presents
real experiments.
\section{Problem Setup: Collective Safety for Multi-robot Surveillance}
\label{sec:problem}
In this section, we introduce the multi-robot problem scenario
addressed in the paper. We start by presenting the communication
model for the cooperating drones.
For each drone $i$, the states and inputs are assumed to belong to a
given set, i.e., $\stateti \in \mathcal{X} \subseteq \real^{\sdim}$
and $\inputti \in \mathcal{U} \subseteq \real^{\udim}$, and to
evolve according to the linear dynamics
\begin{equation}
    \statetpi = A\stateti + B\inputti,
    \label{eq:dynamics}
\end{equation}
with state matrix $A \in \mathbb{R}^{\sdim \times \sdim}$, and input matrix $B \in \mathbb{R}^{\sdim \times \udim}$. 
The drones are associated a safety sphere with radius $\rag \in \real$ and nominal position $\pti \in \real^{\pdim}$.
The drones need to cooperatively track a non-cooperative target (i.e., whose
trajectory is not controlled by the drones) while moving safely. 
The team must adapt its mission (follow the target) to avoid collisions with 
obstacles.
The perception of obstacles is outside the scope of this work. We
refer the reader to \cite{schilling2021vision,ebadi2023present} and for the inter-agent and obstacles
detection in challenging environments.
The goal of the paper is to design a hierarchical
architecture that combines 
distributed optimization,
predictive safety filters and a novel compressing mechanism for
handling the large number of obstacles.

To model the information exchange among the set of cooperating drones $\Ag = \until{\N}$, 
we consider a two-layer communication topology that decouples the mission objectives 
from the safety requirements. The high-level optimization layers relies on a static, 
undirected graph, which is assumed to be connected, with a doubly stochastic adjacency matrix 
$\adj \in \real^{\N \times \N}$ and neighbor set $\Neigha$. Conversely, the low-level safety filter operates 
on a time-varying, distance-based graph. In this layer, the neighborhood of drone $i$ at time 
$\ti$ is defined as $\Neighs = \big\{j \in \Ag \mid \left\|\pti - \ptj\right\| \le \sdist\big\}$, 
where $\sdist$ is the communication threshold.

Let us introduce the multi-robot task addressed
in the paper.
Let $\targett\in\real^{\pdim}$ be the position of the target at time
$\ti$.
Each drone $i \in \Ag$ is aware of the current position of its
neighbors, but does not know their future movements.
We say that the team of drones is \emph{collectively safe} when the
formation satisfies the following definition.
  \begin{definition}[Collective Safety] Let
    $\pti \in \real^{\pdim}$ be the position of drone $i$ at time
    $\ti$. The team is said to be collectively safe when the following
    conditions hold:
  \begin{enumerate}
	\item \textit{Static obstacle avoidance}: Each drone $i \in \Ag$ must avoid collision with the static obstacles located at $\obsj \in \real^{\pdim}$ with $j \in \until{\Nobs}$, where $\Nobs \in \mathbb{N}$ is the number of obstacles. 
	This constraint can be written as: 
	\begin{equation}
	  \left\|\pti - \obsj\right\| \ge \rag + \robs + \delta_{\text{o}},
	\end{equation}
	with $\rag, \robs \in \real$ the radii of the drone and of the obstacle, respectively, and $\delta_{\text{o}} \in \real$ a safety margin.
	\item \textit{Inter-agent collision avoidance}: Safety among drones is guaranteed when the following inequality holds:
	\begin{equation}
	  \left\|\pti - \ptj\right\| \ge 2\rag + \delta_{\text{a}},
	\end{equation}
	for each $j \in \Ag$, $j \neq i$, and $\delta_{\text{a}} \in \real$ a safety margin. \oprocend
  \end{enumerate}
  \label{def:collective_safety}
  \end{definition}
  
  \noindent
  Then, we can state the main task addressed in the paper.
  \begin{problem}[Collectively Safe Target Monitoring]
      \label{prob:task}
      Design a collectively safe distributed motion planning and control strategy that navigates a team of
      drones~\eqref{eq:dynamics} to monitor a moving target
      $\targett$, $\ti \ge 0$.
  \end{problem}
  
\section{Hierarchically-Distributed Safe Surveillance Strategy}
\label{sec:control}
\begin{figure}[]
    \centering
    \includegraphics[width = 0.7\columnwidth]{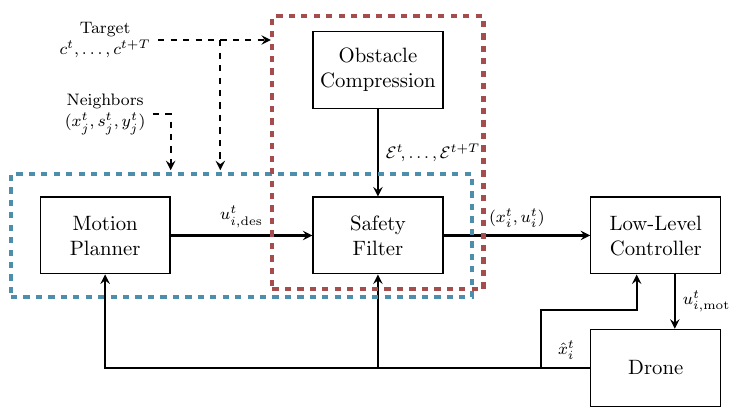}
    \caption{Block diagram of the proposed solution for the drone $i$ at a generic time $\ti$.}
    \label{fig:block}
  \end{figure}

In this section, we present the hierarchically-distributed control
strategy to solve Problem~\ref{prob:task}.
The scheme is composed of three main blocks: \textit{(i)} a
distributed optimization-based control scheme to keep the drones
close to the target, \textit{(ii)} a decentralized predictive safety
filter let the drones satisfy input and state constraints,
and \textit{(iii)} a low-level control strategy to drive the drones to the
desired position.
The general framework of the proposed solution for drone $i \in \Ag$ is illustrated in Fig.~\ref{fig:block}. 
The first block, \textit{Motion Planner} in Fig.~\ref{fig:block},
generates the next desired position $\ptides \in \real^{\pdim}$ for
all drones $i \in \Ag$ by means of a distributed online
optimization scheme inspired from the aggregative
optimization
framework~\cite{li2021distributed2,carnevale2022distributed}.
The second block, \textit{Safety Filter} is
a local predictive safety filter, designed to ensure collective safety
of the formation and actuation limits with a minimum deviation from
the desired control input.
Finally, the third block, \textit{Low-Level Controller}, is a
nonlinear geometric controller that stabilizes the closed-loop system
(see~\cite{mellinger2011minimum}).
Drones only know the target $\targett$ position at time $\ti \ge 0$ and its
planned path in a limited horizon.
In Fig.~\ref{fig:block}, the dashed blue box represents the
distributed safe-motion planner, while the dashed red box represents
the safety block, composed by the centralized obstacle compression and
the local predictive safety filter.
The strategy is summarized in Algorithm~\ref{alg:strategy}, with the main
steps, corresponding to the different blocks of Fig.~\ref{fig:block},
detailed in the next subsections.
\begin{figure}[t]
    \begin{algorithm}[H]
        \caption{\textsc{\strategyname/} -- Solution Strategy}
        \label{alg:strategy}
        \begin{algorithmic}
            \State Each drone $i \in \Ag$ runs:
            \For{$\ti = 1,2, \dots$}
            \State \textbf{Aggregative Motion Planner}
            \Statex \hspace{2em} Run update step~\eqref{eq:zdes_agg} to obtain $\ptides$
            \Statex \hspace{2em} Compute (potentially) unsafe reference $\inputtides$~\eqref{eq:propdes}
            \State \textbf{Predictive Safety Filter}
            \Statex \hspace{2em} Receive safe ellipse~\eqref{eq:ellipseshrink}
            \Statex \hspace{2em} Solve \eqref{alg:min_prob_psf} to obtain safe input $\inputti$
            \State \textbf{Low-Level Controller}
            \Statex \hspace{2em} Inject $\inputti$ as reference to controller~\cite{mellinger2011minimum}
            \EndFor
        \end{algorithmic}
    \end{algorithm}
\end{figure}

\subsection{Distributed Aggregative Planner}
To achieve the target monitoring task, we exploit the distributed
aggregative optimization in~\cite{carnevale2022distributed}. At each
$\ti \ge 0$ the $\N$ drones solve an optimization problem of the form
\begin{align}
    \begin{split}
      \min_{\xa\in \real^{\N \pdim}} \: & \: \sum_{i =1}^\N \costati\left(\xai,\sigma\left(\xa\right)\right).
    \end{split}
    \label{eq:aggregative_problem}
\end{align}
Here, $\xa = \col(\xa_{1},\ldots,\xa_{\N})$, each
$\costati(\xai,\sigma(\xa)): \real^{\pdim} \times \real^{\pdim}
\mapsto \real$ depends on the local optimization variable of drone $i$
and on an \emph{aggregative variable}
$\sigma(\xa) = \sum_{i=1}^\N \frac{\phi_{i}(\xai)}{\N}$,
with $\phi_{i}(\xai):\real^{\pdim}\mapsto \real^{\pdim}$ for each $i \in \Ag$.
In our scenario $\phi_i(\xai) = \xai$, thus $\sigma(\xa)$ represents the team centroid.

Problem~\eqref{eq:aggregative_problem} is solved in a distributed way
by using the \emph{aggregative tracking} algorithm proposed in~\cite{li2021distributed2,carnevale2022distributed}. 
At each iteration $\ti$, each drone $i$ computes $\ptides$ emulating a centralized gradient descent step in a distributed fashion.
Since $\sigma(\xat)$ and
$\sum_{i =1}^\N \nabla_2\costati(\xati,\sigma(\xat))$ are not
available to single drones, they track this global information
through two auxiliary variables $s_i^{\ti} \in \real^{\pdim}$ and
$y_i^{\ti} \in \real^{\pdim}$, respectively. Formally, each drone $i$
initializes its states as $\xai^{0} \in \real^{\pdim}$, $s_i^{0} =
\phi_i(\xai^{0})$, $y_i^{0} = \nabla_2 \costa_i^{0}(\xai^{0},s_i^{0})$
and runs at each $\ti$
\begin{align}
  \notag
  &\ptides = \xati - \alpha(\nabla_1\costati(\xati,s_i^{\ti}) + \phi_i(\xati)y_i^t) \\[1.2ex]\label{eq:zdes_agg}
  & s_i^{\ti+1} = \sum_{j \in \Neigha} a_{ij}s_j^{\ti} + \phi_i(\xatpi) - \phi_i(\xai^{\ti})\\[1.2ex]\notag
  &y_i^{\ti+1} = \sum_{j \in \Neigha} a_{ij}y_j^{\ti} + \nabla_2\costatpi(\xatpi,s_i^{\ti+1}) - \nabla_2\costa^{\ti}_i(\xai^{\ti},s_i^{\ti}),
\end{align}
with $\alpha > 0$ a stepsize.
We refer the reader to \cite{carnevale2022distributed} for the formal proof of 
its convergence properties.
We model the cost function as the sum of three terms, i.e., 
\begin{equation}
    \begin{split}
        \costati(\xai,\sigma(\xa)) = &\gamma_1 \left(\left\|\sigma\left(\xa\right) - \xai\right\|^2 - \da^2 \right)^2
        + \gamma_2\left\|\sigma(\xa) - \targett\right\|^2 + \gamma_3\left\|\xai - \ptihat\right\|^2.
    \end{split}
    \label{eq:cost_aggregative}
\end{equation}
The three terms in \eqref{eq:cost_aggregative} aim, respectively, to maintain the drone at a predetermined distance $\da \in \real$ from the formation centroid $\sigma(\xa)$, 
to guide the centroid 
toward the target current position $\targett$, and to penalize excessive deviations from the drone currently measured position $\ptihat$, 
with $\gamma_1, \gamma_2, \gamma_3 \in \real$ being weights that balance the three terms.
A rigid formation could be enforced by modifying the first term in \eqref{eq:cost_aggregative}.
However, we opted for a loosely weighted distance-based cost.
This avoids enforcing a predefined topology, allowing the swarm to autonomously adapt its shape to safety constraints.
\strategyname/ 
assumes global target state knowledge in~\eqref{eq:cost_aggregative}, however this can be challenging in non-cooperative scenarios. 
Integrating a distributed estimator is under current investigation.
Finally, $\inputtides \in \real^{\pdim}$ is computed as
\begin{align}
    \inputtides = \Pi\left(\ptides,\xati\right),
    \label{eq:propdes}
\end{align}
where $\Pi: \real^{\pdim} \times \real^{\pdim} \to \real^{\udim}$ denotes a control policy designed by the user to steer the drone from $\xati$ to $\ptides$.

\subsection{Obstacle Compression Node}
\label{sec:compression_node}
To ensure collision avoidance without relying on a centralized coordinator, the drones 
cooperatively compute a safe convex region at each time step $\ti$. 
Specifically, each drone $i \in \Ag$ locally computes a safe set 
$\ellipset = \{B_i^\ti r + \cellti \mid \|r\| \le 1\}$ centered at $\cellti \in \real^{\pdim}$ with eccentricity matrix $B_i^\ti \in \real^{\pdim \times \pdim}$.

We consider a scenario where each drone is aware only of a local subset of obstacles. For clarity, assume each drone $i$ tracks a single local obstacle $o_i$ (the extension to multiple obstacles per drone is straightforward). The drone formulates a hyperplane separating the target $\targett$ from the obstacle:
\begin{equation}
    \label{eq:hyperplane}
    a_i^\ti = \frac{o_i - \targett}{\|o_i - \targett\|}, \quad b_i^\ti = (o_i - \robs a_i^\ti)^{\top}a_i^\ti.
\end{equation}
To maximize the safe area, we exploit the well-known maximum volume inscribed ellipsoid problem. Since no drone has global knowledge of all obstacles, we introduce a distributed consensus scheme (Algorithm~\ref{alg:constraintcons}) inspired by~\cite{notarstefano2011distributed}. 
Each drone iteratively solves the optimization problem using a local set of constraints $\activeset_i^\ti$, exchanging only the active hyperplanes with its communication neighbors $j \in \Neighs$:
\begin{align}
    \label{eq:ellipsator}
    B_i^\ti, \cellti = \argmax_{B, c} \quad & \beta_1\log(\text{det}(B)) - \beta_2\|\targett - c\|^2 \\
    \textrm{s.t.} \quad & \|B a_j\| + a_j^\top c \le b_j, \quad \forall (a_j, b_j) \in \activeset_i^\ti \notag \\
    & B \succ 0 \notag
\end{align}
where $\beta_1, \beta_2 > 0$ balance the ellipse area and its distance from the target.

\begin{algorithm}[htbp]
    \caption{\!Distributed Ellipse Consensus (Drone $i$, time $\ti$)}
    \begin{algorithmic}
        \State \textbf{Initialization:} Target position $\targett$, local obstacle $o_i$
        \State Compute local hyperplane $(a_i^\ti, b_i^\ti)$ using \eqref{eq:hyperplane}
        \State $\tactseti^0 \leftarrow \{(a_i^\ti, b_i^\ti)\}$
        \For{$k = 0, \dots, K-1$}
            \State $\tBki, \tcellki \leftarrow$ solve \eqref{eq:ellipsator} with $\targett$ and $\tactsetki$
            \State $\tconstrexcki \leftarrow$ Extract active constraints from the solution
            \State Send $\tconstrexcki$ to neighbors $j \in \Neigha$
            \State Receive $\tconstrexckj$ from neighbors $j \in \Neigha$
            \State $\tactseti^{k+1} \leftarrow \tactseti^{0} \cup \tconstrexcki \cup \left( \bigcup_{j \in \Neigha} \tconstrexckj \right)$
        \EndFor
        \State \textbf{Final assignment:} $\Bti \leftarrow \tBki$, $\cellti \leftarrow \tcellki$, $\activeset_{i}^{\ti} \leftarrow \tactseti^{k+1}$ \\
        \Return $\Bti, \cellti, \activeset_{i}^{\ti}$
    \end{algorithmic}
    \label{alg:constraintcons}
\end{algorithm}

The ellipse is then expressed in quadratic form as:
\begin{equation}
    \label{eq:ellipsequad}
    \ellipset = \left\{r \in \real^\pdim\mid \left(r-\cellti\right)^{\top}\Mmatt\left(r-\cellti\right) = 1\right\},
\end{equation}
where $\Mmatt = (B_i^\ti B_i^\ti)^{-1}$. 
To account for the drone physical volume, we shrink the ellipse proportionally to the drone radius $\rag$, obtaining the safe shape matrix $\Mmattb = \Sigma_i^\ti \odot \Mmatt$, where $\Sigma_i^\ti = (1 - \rag\sqrt{\lambda_{\text{max}}^\ti})^{-2} I_n$, and $\lambda_{\text{max}}^\ti$ is the maximum eigenvalue of $\Mmatt$. The shrunk safe shape is defined as:
\begin{equation}
    \label{eq:ellipseshrink}
    \ellipsetb = \left\{r \in \real^\pdim\mid \left(r-\cellti\right)^{\top}\Mmattb\left(r-\cellti\right) = 1\right\}.
\end{equation}
The safe region constraint for drone $i$ is then $(\pti - \cellti)^{\top}\Mmattb(\pti - \cellti) \le 1$.

Finally, the local predictive safety filter requires the safe area over 
the entire prediction horizon $\iter \in \tHor$.
To limit the communication overhead, drones run Algorithm~\ref{alg:constraintcons} 
only for the current step $\ti$. 
As detailed in Algorithm~\ref{alg:distrellipse}, for the future steps $\iter$, 
drone $i$ locally propagates the final active constraint set $\activeset_i^\ti$ by 
recalculating the hyperplanes~\eqref{eq:hyperplane} relative to the predicted target 
trajectory $\target^\iter$ and solving~\eqref{eq:ellipsator} locally.

\begin{algorithm}
    \caption{Safe Ellipse Horizon Generation (Drone $i$)}
    \label{alg:distrellipse}
    \begin{algorithmic}
        \For{$\ti = 0, 1, \dots$}
            \State Receive target trajectory $\target^{\ti:\ti+\Hor}$
            \State $\Bti, \cellti, \activeseti^{\ti} \leftarrow$ Run \text{Algorithm~\ref{alg:constraintcons}} at time step $\ti$
            \For{$\iter = \ti+1, \dots, \ti+\Hor$}
                \State $\activeseti^{\iter} \leftarrow$ Update hyperplanes in $\activeseti^{\ti}$ with $\target^\iter$
                \State $B_i^\iter, c_{\text{ell}, i}^\iter \leftarrow$ solve \eqref{eq:ellipsator} locally using $\target^\iter$ and $\activeseti^{\iter}$
            \EndFor
        \EndFor
    \end{algorithmic}
\end{algorithm}

\subsection{Safety Filter Block}
\label{sec:filter}
For the safety part, we design a decentralized predictive safety
filter that leverages MPC techniques.
The goal of each drone is threefold, i.e., \textit{(i)} ensure
collective safety at each $\ti$ within the prediction horizon,
\textit{(ii)} satisfy actuation limits, and \textit{(iii)} find
$\inputti$ that gives minimum deviation from $\inputtides$.

Let $\Hor \in \mathbb{N}$ be a prediction horizon, and
$\stateti \in \real^{\sdim}$ and $\inputti \in \real^{\udim}$ the
state and input at time $\ti$, respectively. At each $\ti \ge 0$, each
drone $i \in \Ag$ computes a sequence of states and control inputs
$\statebfi = \col(\stateti,\ldots,\state^{\ti+\Hor}_{i})$,
$\inputbfi = \col(\inputti,\ldots,\uu^{\ti+\Hor}_{i})$ as solution of
the following optimization problem

\begin{subequations}
    \begin{align}
        \min_{\substack{\statebfi,\inputbfi,\slacki}} \quad & \costsuti\left(\inputbfi,\slacki\right) \label{eq:predictive:cost} \\
        \textrm{s.t.} \quad &
        \statekpi = A\stateki + B\inputki \label{eq:pdyn}
        \\ 
        & [ \pki, \velki]^{\top} = C\stateki  \label{eq:poutput}
        \\
        & 1 - \left(\pki - \cellk\right)^{\top}\Mmatkb\left(\pki - \cellk\right) \ge 0 
        \label{eq:constrell}
        \\
        &{\zkj}^{\top}\left(\pki - \mkj - \left(\rag -
          \slackkj\right)\zkj\right) \ge 0
          \label{eq:constr_interagent}
        \\
        &0 \le \slackj^{\tau} \le
          \varepsilon_{\text{max}} \label{eq:slackbound}\\
        &\slackj^{t} = 0, \quad \slackj^{t+1}=0 \qquad j \in \Neighs \label{eq:slackinitial}
        \\
        &\velmin \le \velki \le \velmax, \quad \inputmin \le \inputki \le \inputmax 
        \label{eq:velbound}
        \\ 
        &\stateti = \hat{\state}^t_i \label{eq:pinitial}
        \\
        &\state^{\ti+\Hor}_{i} \in \mathcal{X}^{\Hor}_i
        \notag
        \\
        &\forall \iter \in \tHor
        \notag
    \end{align}
    \label{alg:min_prob_psf}
\end{subequations}
where $\hat{\state}^t_i$ is the current measured state of drone $i$,
and $C \in \real^{2\pdim \times \sdim}$ is the output matrix. The
dynamics and the output equations are imposed in \eqref{eq:pdyn}
and~\eqref{eq:poutput} respectively. Then
constraint~\eqref{eq:constrell} ensures that the drone is inside the
safe ellipse, while~\eqref{eq:pinitial} ensures that the
initialization is consistent with the current state of the drone. The
design of the cost function and of the remaining constraints are detailed in the following.

\begin{remark}
Each drone solves problem~\eqref{alg:min_prob_psf} using only its neighbors current positions, without knowing the desired future trajectory or the MPC solutions of the other drones.

\oprocend
\end{remark}
We consider the cost function
\begin{equation}
    \begin{split}
        \costsuti\left(\inputbfi,\slacki\right) = &\rho_1\left\|\uu_i^{\ti} - \uu_{i,\text{des}}^{\ti}\right\|^2 + \rho_2\sum_{\iter = \ti+1}^{\ti+\Hor}\left\|\inputki\right\|^2
        + \rho_3\sum_{\iter = \ti}^{\ti+\Hor}\sum_{j = 1}^{|\Neighs|-1}\left\|\slackkj\right\|^2,
    \end{split}
\end{equation}
where $\rho_1, \rho_2, \rho_3 \in \real$ are weights that balance the
three terms.
The first and second term represent respectively a penalty on the deviation
from the desired control input and a regularization term. The last
term involves slack variables used to enforce some soft constraints.
Specifically, we introduce a set of slack variables
$\slacki = \col(\slacki^{\ti},\ldots,\slacki^{\ti+\Hor})$ with
$\slacki^{\iter} =
\col(\slack^{\iter}_{i1},\ldots,\slack^{\iter}_{i|\Neighs|-1})$ for
$\iter \in \tHor$, that relax inter-agent collision avoidance
constraints defined as follows.
For each drone $i \in \Ag$, given the set of neighbors at the time
$\ti$, inter-agent constraints are based on hyperplanes that restrict the
available space of the future trajectory.
To this end, let $\ztj\in \real^{\pdim}$ be the unit vector pointing
from drone $i$ to drone $j$ and $\mtj \in \real^{\pdim}$ the midpoint
between drones i.e.,
\begin{align}
 \ztj = \frac{\pti - \ptj}{\left\|\pti - \ptj\right\|}, \quad \mtj = \frac{\pti + \ptj}{2},
\end{align}
respectively.
Then, the constraints between each drone $i$ and its neighbors
$j \in \Neighs$, required to ensure inter-agent collision avoidance at
time $\ti$ read as
\begin{align}
    \label{eq:constr:inter:t:nofoot}
    {\ztj}^{\top}\left(\pti - \mtj\right) \ge 0, \qquad \forall i \in \Ag, j \in \Neighs.
\end{align}
To ensure a proper inter-agent collision avoidance, we rewrite the
constraint~\eqref{eq:constr:inter:t:nofoot} by taking into account also
the footprint of the drones, namely including a circle of radius
$\rag$, as
\begin{align}
    \label{eq:constr:inter:t}
    {\ztj}^{\top}\left(\pti - \mtj - \rag\ztj\right) \ge 0, \qquad \forall i \in \Ag, j \in \Neighs.
\end{align}
Notice that constraint~\eqref{eq:constr:inter:t} defines a separating
hyperplane between each drones and its neighbors at time $\ti$ only.

\begin{figure}[]
\centering
    \includegraphics[width=0.45\columnwidth]{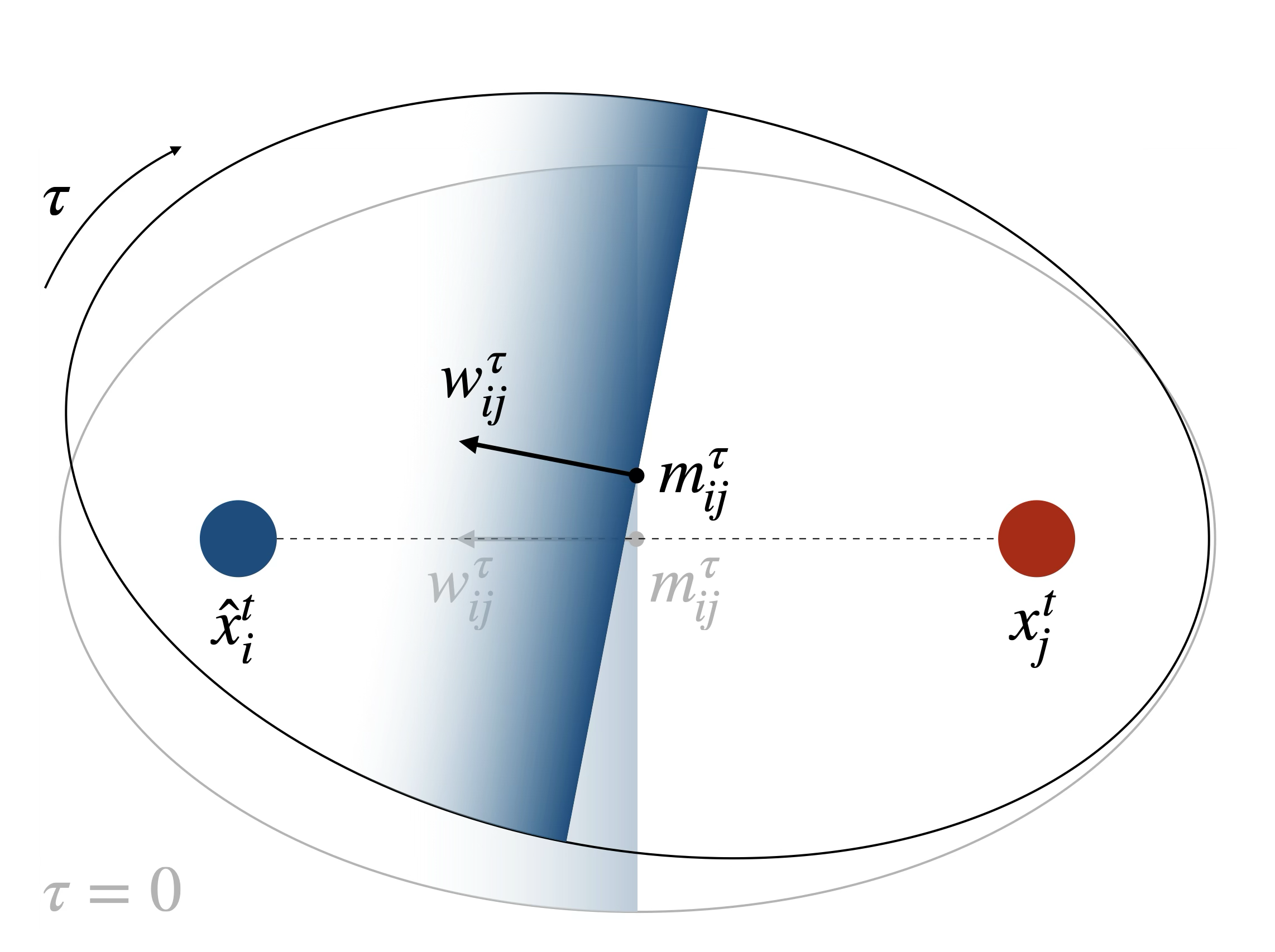}
        \caption{Graphical representation of the inter-agent collision avoidance constraints described in \eqref{eq:constr_interagent}.}
        \label{fig:interagent_ca}
\end{figure}

To enforce safety at the next time instants, the hyperplane
constraints are propagated along the prediction horizon by taking into
account the evolution of the ellipse in~\eqref{eq:ellipseshrink}.
In particular, let $\lambda_i^{\iter}$ and $\mathrm{v}_i^{\iter}$,
$i \in \until{\pdim}$, be, respectively, the eigenvalues and the
associated eigenvectors of matrix $\Mmatkb$
in~\eqref{eq:ellipseshrink}. Let
$D^{\iter} =
\text{diag}_i{\bigl\{({\lambda^{\iter}_i})^{-1/2}\bigr\}}$ be a
transformation which takes into account the eccentricity of the
ellipsoid and
$R^{\iter} =
\bigl\{\mathrm{v}^{\iter}_1,\ldots,\mathrm{v}^{\iter}_{\pdim}\bigr\}$
a rotation matrix that accounts for the orientation thereof. We apply
a linear transformation $\Tk: \real^{n} \to \real^{n}$ defined as $\Tk = \psi^{\iter+1} \circ \left({\psi^{\iter}}\right)^{-1}$,
where $\psi^{\iter} = R^{\iter}D^{\iter}$.
Hence, the update rule for the normal unit vector of the $j$-th hyperplane reads as
\begin{align}
    \zkpj = \Tk \zkj, \quad \iter \in \{\ti,\dots,\ti+\Hor-1\},
\end{align}
while the middle points are translated according to 
\begin{align}
    \mkpj = \mkj + \targetkp - \targetk, \quad \iter \in \{\ti,\dots,\ti+\Hor-1\}.
\end{align}
Next, we introduce a set of slack variables $\slackkj \ge 0$ that
relax the hyperplane constraints for time instants $\tau > t+1$. This
is used to give more degrees of freedom to the optimization solver to
find a viable path.
Each variable $\slackkj$ indicates the amount by which the $j$-th
hyperplane constraint is violated at time $\iter$. For safety
reasons, we impose that the slack variables are bounded by a certain
value $\varepsilon_{\text{max}} \in \real$.
We formalize it by means of the
constraints~\eqref{eq:constr_interagent},~\eqref{eq:slackbound}
and~\eqref{eq:slackinitial}.
We graphically represent the constraints in Fig.~\ref{fig:interagent_ca}.
Notice that, we consider the set of neighbors $\Neighs$ to be fixed
along the entire prediction horizon $\iter \in \tHor$.

We consider drones with actuation limits~\eqref{eq:velbound}, where $\velmin, \velmax \in \real^{\pdim}$ and $\inputmin, \inputmax \in \real^{\udim}$ are the minimum and maximum velocity and acceleration input, respectively.
To ensure feasibility of the optimization problem we impose a terminal constraint that can be divided into two parts, i.e., the final position needs to be inside the safe ellipse and the final velocity needs to be zero.
In particular we define the terminal set $\mathcal{X}^{\Hor}_i$, for all $i \in \Ag$, as
\begin{align}
   \mathcal{X}^{\Hor}_i \!=\! \left\{\state^{\Hor}\!\! \in\! \real^{\sdim} \mid\! \left(\pii^{\Hor} - \cell^{\Hor}\right)^{\top}\!\!\bar{\Mmat}^{\Hor}\left(\pii^{\Hor} - \cell^{\Hor}\right) \le 1, \vel^{\Hor}_i = 0\right\}\!.\notag
\end{align}

\section{Virtual Experiments via Robotic Simulator}
\label{sec:implementation}

We validate the \strategyname/ strategy (Algorithm~\ref{alg:strategy}) through virtual experiments 
in a forest-like scenario, where aerial drones monitor a mobile ground target. 
Simulations are conducted in Webots using \textsc{CrazyChoir}~\cite{pichierri2023crazychoir}, a ROS~2 extension of the \textsc{ChoiRbot} 
toolbox~\cite{testa2021choirbot} for Crazyflie swarms. To ensure modularity and ease real-world deployment, 
each drone is managed by independent ROS~2 processes and communicates only with local neighbors over a sparse graph.

We specialize the desired input \eqref{eq:propdes} as:
\begin{align}
    \inputtides &= K_p\perr + K_v \velerr + K_i e_i^{\ti} \label{eq:propdes:exp}\\
    e_i^{\ti+1} &= e_i^{\ti} + \dt \:\perr, \quad e_i^{0} = 0, \notag 
\end{align} 
where $\dt \in \real$ is the discretization step,
$\perr = \xproj - \ptihat$, $\velerr = \veldes - \veltihat$,
reference velocity $\veldes = [0,0]^\top$, and $K_p, K_v, K_i \in \real^{\udim \times \pdim}$ are feedback 
gain diagonal matrices. Here, $\xproj$ projects $\ptides$ onto the safe set $\ellipsesmall$ 
shrunk as in~\eqref{eq:ellipseshrink}.

\subsection{Multi-Robot Safe Target Tracking}
\label{sec:virtual_experiments}
\begin{figure}[]
\centering
    \includegraphics[width=0.75\columnwidth]{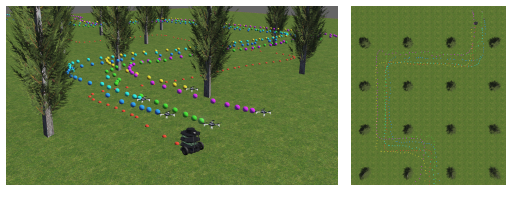}
    \caption{Snapshots from the virtual experiment. A final snapshot
      of the monitoring phase (left) and a top-down view showcasing
      the drones' trajectories inside the forest (right).}
    \label{fig:virtualexperiment}
\end{figure}

\begin{figure}[]
\centering
    \includegraphics[width=0.75\columnwidth]{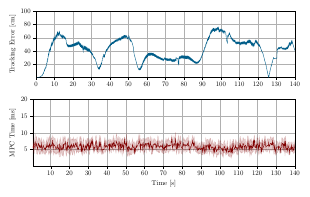}
    \caption{Evolution of the target tracking error (blue) and the computational time (red) of the Predictive Safety Filter.}
    \label{fig:virtualexperiment_performance}
\end{figure}

We consider a 2D scenario, i.e., $\pdim = 2$, where $\N = 5$ Crazyflies track a Turtlebot $3$ Burger 
through a dense matrix of $\Nobs = 16$ trees.
Drone and obstacle radii are $\rag = 0.1 \text{ m}$ and $\robs = 0.15 \text{ m}$, respectively.
We consider a scenario in which all the drones must fly at the same
height, thus making the collision avoidance problem more challenging. 
The drones are modeled as double integrators ($\sdim = 4$, $\udim = 2$) with states 
$\stateti = [\p^{\ti}_{i}, \vel^{\ti}_{i}]^{\top}$ and sampling time
$\dt = 0.1 \text{ s}$, while the ground target follows a pre-computed
cubic spline.
Aggregative planner parameters \eqref{eq:cost_aggregative} are $\gamma_1 = 1.0$, $\gamma_2 = 40.0$, 
$\gamma_3 = 10.0$, desired distance $\da = 0.4\text{ m}$, and stepsize $\alpha = 10^{-3}$. 
Low-level gains \eqref{eq:propdes:exp} are $K_p = 0.4I_2$, $K_v = 1.5I_2$, and $K_i = 0.01I_2$.
Safety filter parameters \eqref{eq:predictive:cost} are $\rho_1 = 20.0$, $\rho_2 = 0.01$, $\rho_3 = 600$, 
and prediction horizon $\Hor = 35$. Ellipse weights \eqref{eq:ellipsator} are $\beta_1 = \beta_2 = 1.0$. 
Actuation limits \eqref{eq:velbound} are bounded at $\pm 1.5 \text{ m/s}$ for velocity and $\pm 1.5 \text{ m/s}^2$ 
for acceleration. The sensing range is $\sdist = 1.0 \text{ m}$.
Simulations are executed on a workstation equipped with an Intel i9 
processor and an Nvidia RTX 4000 GPU. The local optimization problem is formulated 
via CVXPY and solved using CLARABEL~\cite{goulart2026clarabel}. 
The safety filter operates at a frequency of 7 Hz.
This choice is motivated by the need to maintain a real-time reactive control loop: obstacle avoidance algorithms typically 
operate at a few Hertz to tens of Hertz, where this range is sufficient to repeatedly sense–decide–act and react to environmental changes 
without incurring excessive computational overhead. We experimentally verified that higher frequencies did not provide significant performance improvements.
\begin{figure*}[]
\centering
    \includegraphics[width=\textwidth]{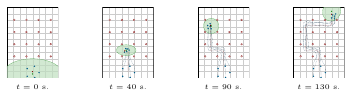}
    \caption{Virtual experiment screenshots at $0,40,90,130$ $[$s$]$. Red circles are obstacles, the red dot is the target, and blue dots are drones. Dashed lines are the trajectories up to $\ti$.}
        \label{fig:numsimulation}
\end{figure*}
The mobile ground robot follows a pre-computed cubic spline via the
Unicycle control suite of \textsc{ChoiRbot}. The Crazyflies, instead,
are controlled by the proposed hierarchically-distributed predictive
control strategy deployed on \textsc{CrazyChoir}.
Fig.~\ref{fig:numsimulation} illustrates the experiment evolution at $t \in \{0,40,90,130\} \text{ s}$, 
highlighting the ellipse compression adapting to different forest zones. Fig.~\ref{fig:virtualexperiment} 
provides a final snapshot and a top-down trajectory view, demonstrating safe target monitoring in a cluttered environment.
As reported in Fig.~\ref{fig:virtualexperiment_performance}, 
the tracking error remains below $0.8 \text{ m}$ throughout the mission. 
The Predictive Safety Filter executes in $6 \text{ ms}$ on average,
validating its real-time capability.
Furthermore, the maximum worst-case execution time peaks at strictly 11.6 ms.
The distributed
optimization step \eqref{eq:zdes_agg}, implemented via the
\textsc{ChoiRbot} toolbox \cite{testa2021choirbot} (leveraging
DISROPT), averages just $0.01 \text{ ms}$, yielding a negligible
computational footprint compared to the downstream MPC.

\begin{figure*}[]
    \centering
    \includegraphics[width=\textwidth]{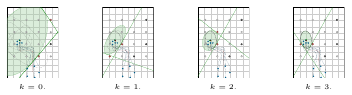}
    \caption{Evolution of Distributed Ellipse Consensus
      (Algorithm~\ref{alg:constraintcons}) for Drone 5 at fixed time $\ti$. From the local initialization ($k=0$) to consensus
      ($k=3$). Active separating hyperplanes (green lines) are
      exchanged among neighbors. Red dots represent active
      constraints, while dark gray ones are locally known but
      inactive, and dashed lines indicate the ellipses computed by the
      other drones.}
    \label{fig:ellipse_consensus}
\end{figure*}

To illustrate the distributed ellipse consensus
(Algorithm~\ref{alg:constraintcons}), Fig.~\ref{fig:ellipse_consensus}
details the local safe set computation for a single drone (e.g., Drone
5) at a generic time step $\ti$. At $k=0$,
the drone relies solely on its locally sensed obstacles, formulating local separating hyperplanes. 
Throughout Algorithm~\ref{alg:constraintcons},
each drone exchanges active constraints with its neighbors. Setting
the maximum iterations to $K=\N=5$, the local ellipse rapidly
converges to the globally safe configuration in just $k=3$ steps. This
confirms that the decentralized scheme effectively aggregates the
scattered environmental geometry without a central coordinator.

\begin{figure}[]
\centering
    \includegraphics[width=0.75\columnwidth]{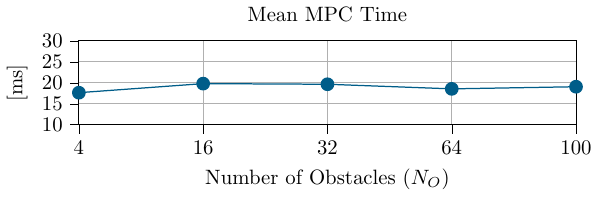}
    \caption{Computation time of the \strategyname/ Predictive Safety Filter increasing the number of obstacles $\Nobs$.}
    \label{fig:virtualexperiment_compression}
\end{figure}

To evaluate the constraint compression performance, we repeated the
experiment by varying the obstacle density within the same workspace
($\Nobs \in \{4, 16, 32, 64, 100\}$). As shown in
Fig.~\ref{fig:virtualexperiment_compression}, the Predictive Safety
Filter computation time remains almost constant, highlighting the
effectiveness of the compression mechanism against environment
clutter.

We point out that while in this virtual experiment the fully
distributed ellipse consensus (Algorithm~\ref{alg:constraintcons}) is
run, the subsequent Monte Carlo campaigns and real-world experiments
employ a parallel architecture with a single obstacle compression node (see Section~\ref{sec:compression_node}). This
choice is motivated by the need to mitigate communication latency and
computational overhead during extensive testing, thereby allowing the
performance analysis to focus strictly on the decentralized
scalability and real-time reliability of the local Predictive Safety
Filters.

\subsection{Monte Carlo Virtual Experiments}
\label{subsec:monte_carlo}
To evaluate Algorithm~\ref{alg:strategy}, we run two Monte Carlo campaigns analyzing performance 
against varying fleet sizes $\N$ and target velocities $\targetvel$. 
Scenarios are generalized by randomizing obstacle and target positions, initial drone formation, and communication graphs. 
In each trial, the target follows an A$^\star$-generated trajectory at a mean velocity $\targetvel$.

\begin{figure}[]
    \centering
    \includegraphics[width=0.65\columnwidth]{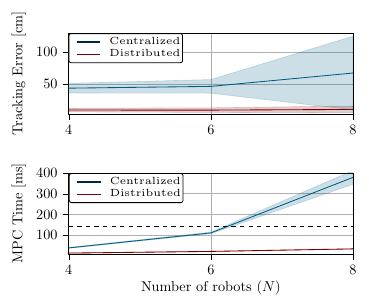}
    \caption{Algorithm performance increasing the number of drones
      $\N$. Target tracking error (above) and computation time of the
      optimal solution (below) are shown. The shaded areas represent the
      standard deviation band, while the dashed black line indicates
      the $\approx 142 \text{ ms}$ real-time deadline (corresponding
      to the $7 \text{ Hz}$ safety frequency). In blue the centralized filter, in red our distributed approach.}
    \label{fig:montecarlo_1}
\end{figure}

First, we evaluate scalability ($\N \in \{4, 6, 8\}$, $50$ trials, $\targetvel=0.08 \text{ m/s}$) against a centralized 
PSF baseline adapted from~\cite{safaoui2024safe}. In this centralized setup, instead of using distributed local solvers, a 
single MPC gathers all desired inputs to jointly enforce inter-agent collisions and compute all safe commands.
As shown in Fig.~\ref{fig:montecarlo_1}, the centralized approach (blue line) exhibits exponential computational growth. 
At $\N=8$, it takes nearly $400 \text{ ms}$, severely violating the $142 \text{ ms}$ real-time deadline ($7 \text{ Hz}$, dashed line). 
Even when meeting the deadline (e.g., at $\N=4$), the centralized solver yields a higher average tracking 
error ($\approx 50 \text{ cm}$). This performance drop is rooted in
the monolithic nature of the baseline.
Furthermore, centrally resolving all inter-agent constraints simultaneously induces chattering, 
further deviating the swarm from the reference path.
Conversely, our strategy (red line) effectively distributes the computational effort, keeping the tracking error tightly 
bounded around $10 \text{ cm}$. Since each drone solves a localized, fixed-dimension problem~\eqref{alg:min_prob_psf}, execution times 
increase marginally, staying safely below $40 \text{ ms}$ even for
$\N=8$.

\begin{figure}[!b]
    \centering
    \includegraphics[width=0.45\columnwidth]{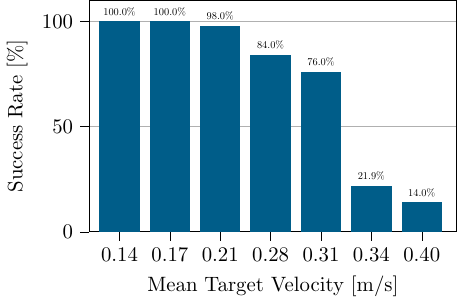}
    \caption{
        Mission success rate at increasing target velocity $\targetvel$.
    }
    \label{fig:montecarlo_2}
\end{figure}

Second, we evaluate collective safety
(Definition~\ref{def:collective_safety}) by varying the target
velocity $\targetvel$ for a fleet of $\N = 5$ over $50$ trials per
velocity. A mission succeeds if the entire team tracks the target
without any collisions.
As shown in Fig.~\ref{fig:montecarlo_2}, the success rate approaches
$100\%$ for velocities $\targetvel \le 0.21 \text{ m/s}$, confirming
the predictive filter's reliability. The expected performance drop at
higher speeds is inherent to the distributed architecture: lacking
global knowledge of neighbors' future trajectories, tracking a fast
target in clutter severely restricts the local MPC feasible space.

\section{Real-World Experiment: Nano-Quadrotor Swarm Monitoring a Mobile Ground Robot}
\label{sec:experiments}
\begin{figure}[htbp]
\centering
    \includegraphics[width=0.75\columnwidth]{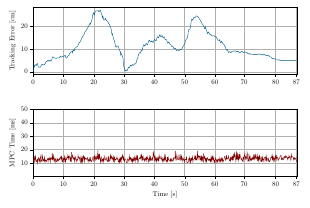}
    \caption{Evolution of the target tracking error (blue) and the computational time (red) of the Predictive Safety Filter.}
    \label{fig:realexperiment_performance}
\end{figure}

\begin{figure}[htbp]
    \centering
    \includegraphics[width=0.75\columnwidth]{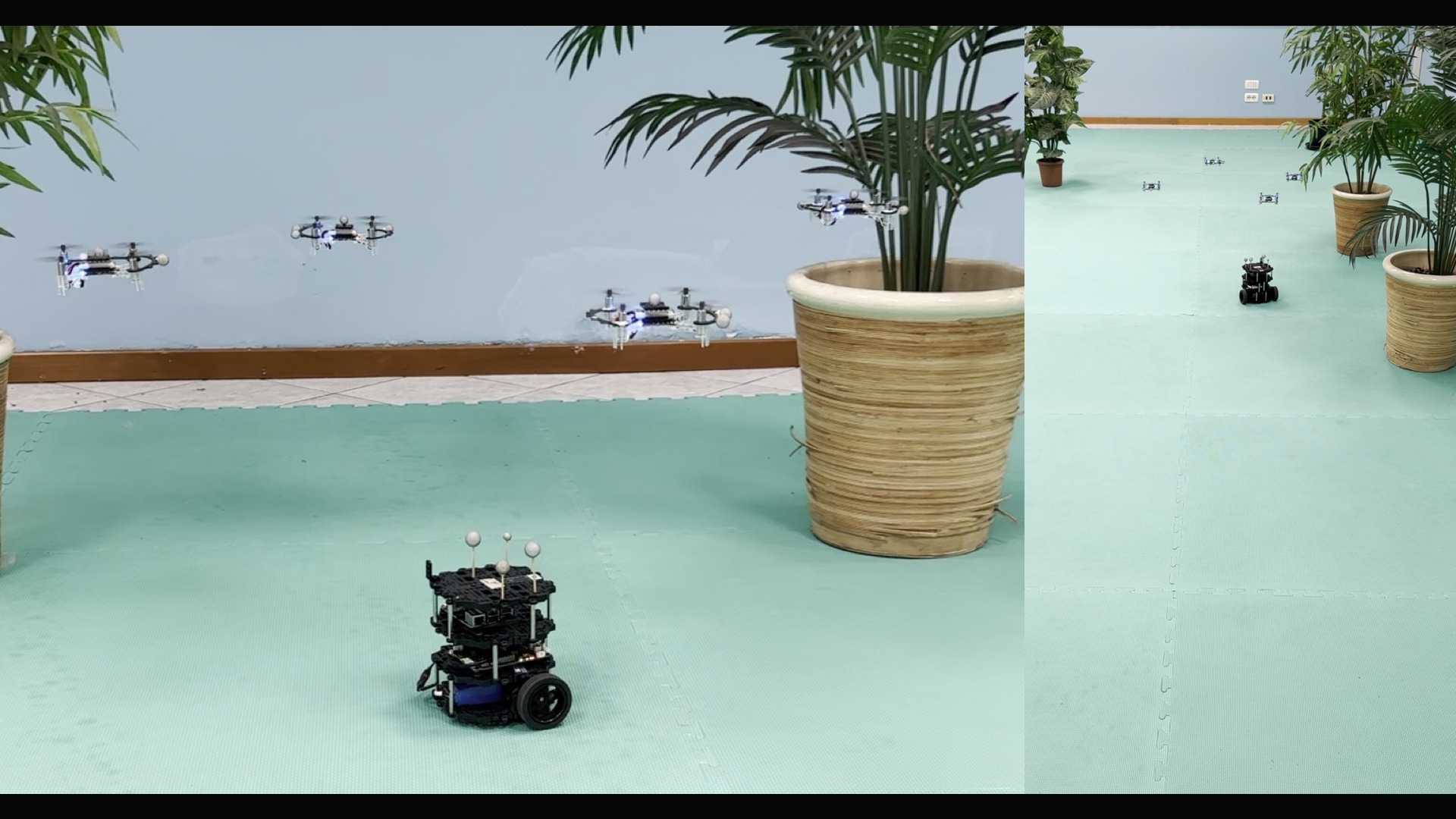}
    \caption{
        Snapshot of the real experiment.
    }
    \label{fig:experiment}
\end{figure}

We validate the proposed \strategyname/ strategy in a real-world
indoor flight arena ($9 \times 4 \times 3$ m) using a fleet of $\N=4$
Crazyflie 2.1 nano-quadrotors and a TurtleBot 3 Burger as the moving
target. Four physical static obstacles are placed in the workspace to
evaluate the collision avoidance performance. Robots states are tracked
at $100$~Hz by a Vicon motion capture system.
The hardware workstation, ROS 2 middleware, and optimization solvers 
employed in this deployment are identical to those detailed in Section IV. 
Due to the computational constraints of the nano-quadrotors, onboard
optimization is infeasible. However, to rigorously evaluate the
distributed nature of \strategyname/ and subject the system to
realistic middleware communication delays, we adopt a decentralized
software architecture. Each drone is governed by independent ROS~2
nodes that do not share memory, exchanging information strictly via
standard network topics to emulate a true distributed computing
environment.
Optimal control inputs are converted
into angular-rate setpoints by the \textsc{CrazyChoir} geometric
controller and transmitted to the Crazyflies at $100$~Hz via a
Crazyradio 2.0 dongle. Concurrently, the TurtleBot 3 receives velocity
commands through the \textsc{ChoiRbot} suite.
To replicate this experiment, parameters are set as follows:
$\dt = 0.14 \text{ s}$, $\gamma_1 = 10.0$, $\gamma_2 = 1.0$,
$\gamma_3 = 0.1$, $\da = 0.5\text{ m}$, $\alpha = 10^{-3}$,
$K_p = 3.75I_2$, $K_v = 4.0I_2$, $K_i = 0.3I_2$, $\rho_1 = 64.0$,
$\rho_2 = 10^{-3}$, $\rho_3 = 75$, $\Hor = 30$, and
$\sdist = 0.8 \text{ m}$. Actuation limits are bounded at
$\pm 1.0 \text{ m/s}$ for velocity and $\pm 1.5 \text{ m/s}^2$ for
acceleration.
As shown in Fig.~\ref{fig:realexperiment_performance}, the real-world
performance closely matches the virtual experiments. The target
tracking error remains strictly bounded, and the computational time of
the Predictive Safety Filter stays within the real-time
deadline. Ultimately, the entire team successfully achieves
the monitoring task while avoiding both inter-agent and static obstacle collisions.
In Fig.~\ref{fig:experiment}, we show a snapshot of the real
experiment, where the Crazyflie swarm monitors the TurtleBot~3 Burger
while ensuring collective safety.

\section{Conclusion}
We proposed \strategyname/, a hierarchically-distributed predictive control strategy 
for cooperative drones navigating obstacle-dense environments. 
By decoupling the high-level distributed aggregative tracking from the low-level 
MPC safety filtering, we ensured both task execution and collective safety. 
Furthermore, a novel distributed obstacle compression mechanism bounds the 
drones motion within a safe elliptical region, overcoming the computational bottlenecks 
of traditional dense-environment navigation. Extensive Monte Carlo simulations 
and real-world experiments validated the strategy,
demonstrating 
superior scalability and real-time computational efficiency compared to centralized 
approaches.
Future work includes the development of a comprehensive theoretical analysis to formally 
guarantee the recursive feasibility of \strategyname/. 
Additionally, distributed estimation schemes can be integrated to handle non-cooperative 
target tracking without relying on global information.

\bibliography{biblio.bib}

@article{saska2016swarm,
  title={Swarm distribution and deployment for cooperative surveillance by micro-aerial vehicles},
  author={Saska, Martin and Von{\'a}sek, Vojt{\v{e}}ch and Chudoba, Jan and Thomas, Justin and Loianno, Giuseppe and Kumar, Vijay},
  journal={Journal of Intelligent \& Robotic Systems},
  volume={84},
  pages={469--492},
  year={2016},
  publisher={Springer}
}

@article{testa2021choirbot,
  title={Choi{R}bot: A {ROS} 2 toolbox for cooperative robotics},
  author={Testa, Andrea and Camisa, Andrea and Notarstefano, Giuseppe},
  journal={IEEE Robotics and Automation Letters},
  volume={6},
  number={2},
  pages={2714--2720},
  year={2021},
  publisher={IEEE}
}

@article{gao22swarm,
author = {Xin Zhou  and Xiangyong Wen  and Zhepei Wang  and Yuman Gao  and Haojia Li  and Qianhao Wang  and Tiankai Yang  and Haojian Lu  and Yanjun Cao  and Chao Xu  and Fei Gao },
title = {Swarm of micro flying robots in the wild},
journal = {Science Robotics},
volume = {7},
number = {66},
pages = {eabm5954},
year = {2022},
doi = {10.1126/scirobotics.abm5954},
}

@article{wang2023distributed,
  title={Distributed relative localization algorithms for multi-robot networks: A survey},
  author={Wang, Shuo and Wang, Yongcai and Li, Deying and Zhao, Qianchuan},
  journal={Sensors},
  volume={23},
  number={5},
  pages={2399},
  year={2023},
  publisher={MDPI}
}

@article{shorinwa2024distributed,
  title={Distributed optimization methods for multi-robot systems: Part 2—A survey},
  author={Shorinwa, Ola and Halsted, Trevor and Yu, Javier and Schwager, Mac},
  journal={IEEE Robotics \& Automation Magazine},
  year={2024},
  publisher={IEEE}
}

@article{testa2025tutorial,
  title={A tutorial on distributed optimization for cooperative robotics: from setups and algorithms to toolboxes and research directions},
  author={Testa, Andrea and Carnevale, Guido and Notarstefano, Giuseppe},
  journal={Proceedings of the IEEE},
  year={2025},
  publisher={IEEE}
}

@article{li2021distributed2,
  title={Distributed online convex optimization with an aggregative variable},
  author={Li, Xiuxian and Yi, Xinlei and Xie, Lihua},
  journal={IEEE Trans. on Control of Network Systems},
  volume={9},
  number={1},
  pages={438--449},
  year={2021},
  publisher={IEEE}
}

@article{carnevale2022distributed,
  title={Distributed online aggregative optimization for dynamic multirobot coordination},
  author={Carnevale, Guido and Camisa, Andrea and Notarstefano, Giuseppe},
  journal={IEEE Trans. on Automatic Control},
  volume={68},
  number={6},
  pages={3736--3743},
  year={2022},
  publisher={IEEE}
}

@article{pichierri2026multi,
  title={Multi-robot target monitoring and encirclement via triggered distributed feedback optimization},
  author={Pichierri, Lorenzo and Carnevale, Guido and Sforni, Lorenzo and Notarstefano, Giuseppe},
  journal={IEEE Trans. on Robotics},
  year={2026},
  publisher={IEEE}
}

@article{huang2025distributed,
  title={Distributed Aggregative Optimization Algorithm for Solving Multi-Robot Formation Problem},
  author={Huang, Jingyi and Yang, Chuanhai and Wu, Shuang and Liu, Qingshan},
  journal={IEEE Trans. on Control of Network Systems},
  year={2025},
  publisher={IEEE}
}

@article{soria2021distributed,
  title={Distributed predictive drone swarms in cluttered environments},
  author={Soria, Enrica and Schiano, Fabrizio and Floreano, Dario},
  journal={IEEE Robotics and Automation Letters},
  volume={7},
  number={1},
  pages={73--80},
  year={2021},
  publisher={IEEE}
}

@article{toumieh2022decentralized,
  title={Decentralized multi-agent planning using model predictive control and time-aware safe corridors},
  author={Toumieh, Charbel and Lambert, Alain},
  journal={IEEE Robotics and Automation Letters},
  volume={7},
  number={4},
  pages={11110--11117},
  year={2022},
  publisher={IEEE}
}

@article{luis2020online,
  title={Online trajectory generation with distributed model predictive control for multi-robot motion planning},
  author={Luis, Carlos E and Vukosavljev, Marijan and Schoellig, Angela P},
  journal={IEEE Robotics and Automation Letters},
  volume={5},
  number={2},
  pages={604--611},
  year={2020},
  publisher={IEEE}
}

@article{zhou2017fast,
  title={Fast, on-line collision avoidance for dynamic vehicles using buffered voronoi cells},
  author={Zhou, Dingjiang and Wang, Zijian and Bandyopadhyay, Saptarshi and Schwager, Mac},
  journal={IEEE Robotics and Automation Letters},
  volume={2},
  number={2},
  pages={1047--1054},
  year={2017},
  publisher={IEEE}
}

@article{vinod2024decentralized,
  title={Decentralized, Safe, Multiagent Motion Planning for Drones Under Uncertainty via Filtered Reinforcement Learning},
  author={Vinod, Abraham P and Safaoui, Sleiman and Summers, Tyler H and Yoshikawa, Nobuyuki and Di Cairano, Stefano},
  journal={IEEE Trans. on Control Systems Technology},
  year={2024},
  publisher={IEEE}
}

@article{schilling2021vision,
  title={Vision-based drone flocking in outdoor environments},
  author={Schilling, Fabian and Schiano, Fabrizio and Floreano, Dario},
  journal={IEEE Robotics and Automation Letters},
  volume={6},
  number={2},
  pages={2954--2961},
  year={2021},
  publisher={IEEE}
}

@article{ebadi2023present,
  title={Present and future of slam in extreme environments: The darpa subt challenge},
  author={Ebadi, Kamak and Bernreiter, Lukas and Biggie, Harel and Catt, Gavin and Chang, Yun and Chatterjee, Arghya and Denniston, Christopher E and Desch{\^e}nes, Simon-Pierre and Harlow, Kyle and Khattak, Shehryar and others},
  journal={IEEE Trans. on Robotics},
  volume={40},
  pages={936--959},
  year={2023},
  publisher={IEEE}
}

@inproceedings{wabersich2018linear,
  title={Linear model predictive safety certification for learning-based control},
  author={Wabersich, Kim P and Zeilinger, Melanie N},
  booktitle={2018 IEEE Conference on Decision and Control (CDC)},
  pages={7130--7135},
  year={2018},
  organization={IEEE}
}

@article{tearle2021predictive,
  title={A predictive safety filter for learning-based racing control},
  author={Tearle, Ben and Wabersich, Kim P and Carron, Andrea and Zeilinger, Melanie N},
  journal={IEEE Robotics and Automation Letters},
  volume={6},
  number={4},
  pages={7635--7642},
  year={2021},
  publisher={IEEE}
}

@inproceedings{bejarano2023multi,
  title={Multi-step model predictive safety filters: Reducing chattering by increasing the prediction horizon},
  author={Bejarano, Federico Pizarro and Brunke, Lukas and Schoellig, Angela P},
  booktitle={2023 62nd IEEE Conference on Decision and Control (CDC)},
  pages={4723--4730},
  year={2023},
  organization={IEEE}
}

@article{vaaler2024modular,
  title={Modular control architecture for safe marine navigation: Reinforcement learning with predictive safety filters},
  author={Vaaler, Aksel and Husa, Svein Jostein and Menges, Daniel and Larsen, Thomas Nakken and Rasheed, Adil},
  journal={Artificial Intelligence},
  volume={336},
  pages={104201},
  year={2024},
  publisher={Elsevier}
}

@article{safaoui2024safe,
  title={Safe multi-agent motion planning under uncertainty for drones using filtered reinforcement learning},
  author={Safaoui, Sleiman and Vinod, Abraham P and Chakrabarty, Ankush and Quirynen, Rien and Yoshikawa, Nobuyuki and Di Cairano, Stefano},
  journal={IEEE Trans. on Robotics},
  year={2024},
  publisher={IEEE}
}

@inproceedings{mellinger2011minimum,
  title={Minimum snap trajectory generation and control for quadrotors},
  author={Mellinger, Daniel and Kumar, Vijay},
  booktitle={2011 IEEE international conference on robotics and automation},
  pages={2520--2525},
  year={2011},
  organization={IEEE}
}

@article{pichierri2023crazychoir,
  title={Crazychoir: Flying swarms of crazyflie quadrotors in ros 2},
  author={Pichierri, Lorenzo and Testa, Andrea and Notarstefano, Giuseppe},
  journal={IEEE Robotics and Automation Letters},
  volume={8},
  number={8},
  pages={4713--4720},
  year={2023},
  publisher={IEEE}
}

@article{notarstefano2011distributed,
  title={Distributed abstract optimization via constraints consensus: Theory and applications},
  author={Notarstefano, Giuseppe and Bullo, Francesco},
  journal={IEEE Transactions on Automatic Control},
  volume={56},
  number={10},
  pages={2247--2261},
  year={2011},
  publisher={IEEE}
}

@article{cavorsi2023multirobot,
  title={Multirobot adversarial resilience using control barrier functions},
  author={Cavorsi, Matthew and Sabattini, Lorenzo and Gil, Stephanie},
  journal={IEEE Transactions on Robotics},
  volume={40},
  pages={797--815},
  year={2023},
  publisher={IEEE}
}

@article{hu2022decentralized,
  title={Decentralized robust collision-avoidance for cooperative multirobot systems: A Gaussian process-based control barrier function approach},
  author={Hu, Yifan and Fu, Junjie and Wen, Guanghui},
  journal={IEEE Transactions on Control of Network Systems},
  volume={10},
  number={2},
  pages={706--717},
  year={2022},
  publisher={IEEE}
}

@inproceedings{cheng2020safe,
  title={Safe multi-agent interaction through robust control barrier functions with learned uncertainties},
  author={Cheng, Richard and Khojasteh, Mohammad Javad and Ames, Aaron D and Burdick, Joel W},
  booktitle={2020 59th IEEE Conference on Decision and Control (CDC)},
  pages={777--783},
  year={2020},
  organization={IEEE}
}

@inproceedings{grover2020deadlock,
	title={Deadlock analysis and resolution for multi-robot systems},
	author={Grover, Jaskaran Singh and Liu, Changliu and Sycara, Katia},
	booktitle={International Workshop on the Algorithmic Foundations of Robotics},
	pages={294--312},
	year={2020},
	organization={Springer}
}

@article{goulart2026clarabel,
  title={Clarabel: An interior-point solver for conic programs with quadratic objectives},
  author={Goulart, Paul J and Chen, Yuwen},
  journal={Mathematical Programming Computation},
  pages={1--83},
  year={2026},
  publisher={Springer}
}
\bibliographystyle{IEEEtran}

\end{document}